\documentclass{river-journal}
\usepackage{rivps}

\usepackage{graphicx}
\usepackage{graphics}
\usepackage{cite}
\usepackage{epsfig,subfigure}
\usepackage{epstopdf}
\usepackage[section]{algorithm}
\usepackage{algorithmic}
\usepackage{wrapfig}

\par

\begin{document}
\begin{opening}
\title{Structure Preserving Large Imagery Reconstruction}
\author{Ju Shen$^1$, Jianjun Yang$^2$, Sami  Taha-abusneineh$^3$, Bryson Payne$^4$, Markus Hitz$^5$}
\institute{$^1$ Department of Computer Science, University of Dayton, 300 College Park, Dayton, OH 45469, USA, \texttt{jshen1@udayton.edu}\\
$^2$ Department of Computer Science and Information Systems, University of North Georgia, Oakwood, GA 30566, USA, \texttt{jianjun.yang@ung.edu}\\
$^3$ Computer Science Department, Palestine Polytechnic University (PPU), Ein Sara Street, Hebron, Palestine, \texttt{staha77@yahoo.com}\\
$^4$ Department of Computer Science and Information Systems, University of North Georgia, Dahlonega, GA 30597, USA, \texttt{ bryson.payne@ung.edu}\\
$^5$ Department of Computer Science and Information Systems, University of North Georgia, Dahlonega, GA 30597, USA, \texttt{markus.hitz@ung.edu}}
\end{opening}

\runningtitle{Sample Document}
\subsection*{Abstract}
With the explosive growth of web-based cameras and mobile devices, billions of photographs are uploaded to the internet. We can trivially collect a huge number of photo streams for various goals, such as image clustering, 3D scene reconstruction, and other big data applications.  However, such tasks are not easy due to the fact the retrieved photos can have large variations in their view perspectives, resolutions, lighting, noises, and  distortions. Furthermore, with the occlusion of unexpected objects like people, vehicles, it is even more challenging to find feature correspondences and reconstruct realistic scenes. In this paper, we propose a structure-based image completion algorithm for object removal that produces visually plausible content with consistent structure and scene texture. We use an edge matching technique to infer the potential structure of the unknown region. Driven by the estimated structure, texture synthesis is performed automatically along the estimated curves. We evaluate the proposed method on different types of images: from highly structured indoor environment to natural scenes. Our experimental results demonstrate satisfactory performance that can be potentially used for subsequent big data processing, such as image localization, object retrieval, and scene reconstruction. Our experiments show that this approach achieves favorable results that outperform existing state-of-the-art techniques.

\section{Introduction}
\label{sec:intro}

In the past few years, the massive collections of imagery on the Internet have inspired a wave of work on many interesting big data topics. For example, by entering a keyword, one can easily download a huge number of photo streams related to it. Moreover, with the recent advance in image processing techniques, such as feature descriptors \cite{wang2013joint}, pixel-domain matrix factorization approaches \cite{wang2013multiple, sunQ2012, wang2012adaptive} or probabilistic optimization \cite{6619001}, images can be read in an automatic manner rather than relying on the associated text. This leads to a revolutionary impact to a broad range of applications, from image clustering or recognition \cite{zhou2010region, 6607541, zhou2013adaptive, li2014graph, sunQ2010, li2014doc, sunQ2014} to video synthesis or reconstruction \cite{6011997, 6379415, 6619002} to cybersecurity via online images analysis \cite{cross13, yang2010,  web14, yang2013} to other scientific applications \cite{wang2014saga, ZhangHong2014, wang2012scimate, ZhangHong2013, ZhangHong2009}.

However, despite the numerous applications, poor accuracy can be yielded due to the large variation of the photo streams, such as resolution, illumination, or photo distortion. In particular, difficulties arise when unexpected objects present on the images. Taking the Google street view as an example, the passing vehicles or walking passengers could affect the accuracy of image matching. Furthermore such unwanted objects also introduce noticeable artifacts and privacy issue in the reconstructed views.

To resolve these issues, object removal \cite{M_00, Criminisi_04} is an effective technique that has been widely used in many fields. A common approach is to use texture synthesis to infer missing pixels in the unknown region, such as \cite{Ashikhmin_01, Efros_01}. Efros and Leung  \cite{Efros_99} use a one-pass greedy algorithm to infer the unknown pixels based on an assumption that the probability distribution of a target pixel's brightness is independent from the rest of the image given its spatial neighborhood. Some studies propose example-based approaches to fill the unknown regions, such as \cite{Efros_01, Efros_99, Ashikhmin_01}. These approaches failed to preserve the potential structures in the unknown region.  Bertalmio \emph{et al}. \cite{Bertalmio_03} apply \emph{partial differential equations} (PDE) to propagate image Laplacians. While the potential structures are improved in the filled region, it suffers from blurred synthesized texture. Drori \emph{et al}. \cite{Drori_03}, propose an enhanced algorithm to improve the rendered texture. Jia \emph{et al}. \cite{Jia_03} propose an texture-segmentation based approach using tensor-voting to achieve the same goal. But their approaches are computationally expensive. A widely used image in-painting technique developed by Criminisi \emph{et al}. \cite{Criminisi_04} aims to fill the missing region by a sequence of ordered patches by using the proposed confidence map. The priority of each patch is determined by the edge strength from the surrounding region. However, the potential structures in the in-painted region can not be well preserved, especially for those images with salient structures.  The authors Sun \emph{et al.} in \cite{sun_05} make an improvement through structure propagation, while this approach requires additional user intervention and the results may depend on the individual animators.

 As an extension of our early work \cite{shen2014completion}, we propose an automatic object removal algorithm for scene completion, which would benefit large imagery processing. The cue of our method is based on the structure and texture consistency. First, it predicts the underlying structure of the occluded region by edge detection and contour analysis. Then structure propagation is applied to the region followed by a patch-based texture synthesis. Our proposed approach has two major contributions. First, given an image and its target region, we develop an automatic curve estimation approach to infer the potential structure.  Second, an orientated patch matching algorithm is designed for texture propagation. Our experiments demonstrate satisfactory results that outperform other techniques in the literature.

The rest of the paper is organized as follows: in section 2, we give an example to demonstrate the basic steps of our image completion algorithm. Then we define the model and notations in section 3. Details are further explained in section 4. The experiment results are presented in section 5. Finally we conclude the paper and our future work in section 6.

\section{A Simple Example}
\label{sec:mot}

The process of our framework is: for a given image, users specify the object for removal by drawing a closed contour around it. The enclosure is considered as the unknown or target region that needs to be filled by the remaining region of the image. Figure a shows an example: the red car is selected as the removing object. In the resulting image, the occluded region is automatically recovered based on the surrounding available pixels.

Our algorithm is based on two observations: spacial texture coherence and structure consistency along the boundaries between the target and source regions. To ensure spacial coherence, many exemplar-based methods have been proposed to find the potential source texture for the target region. By traversing the available pixels from the known region, a set of ``best patches'' are found to fill the target region. Here the definition of ``best patch'' refers to a small region of contiguous pixels from the source region that can maximize a certain spacial coherence constraint specified by different algorithms. A typical example can be found in \cite{Criminisi_04}. However, a naive copy-and-paste of image patches may introduce noticeable artifacts, though the candidate patches can maximize a local coherence. To resolve this problem, structure preservation is considered to ensure the global consistency. There have been several techniques presented for structure propagation to ensure smooth and natural transitions among salient edges , such as the Sun's method \cite{sun_05}, which requires additional user input to finish the task.

\section{The Approach}
\label{sec:def}

First let us define some notations for the rest of paper. The target region for filling is denoted as $\mathbf{\Omega}$; the remaining part $\Phi (= \mathbf{I} - \mathbf{\Omega})$ is the region whose pixels are known. The boundary contours are denoted as $\mathbf{\partial\Omega}$ that separate $\Phi$ and $\mathbf{\Omega}$. A pixel's value is represented by $p = \mathbf{I(x, y)}$, where x and y are the coordinates on the image $\mathbf{I}$. The surrounding neighborhood centered at $(x, y)$ is considered as a patch, denoted by $\Psi_p$, whose coordinates are within $[x\pm \Delta{x}, y\pm \Delta{y}]$. In our framework, there are three stages involved:structure estimation, structure propagation, and remaining part filling.

{\bf Structure Estimation:} In this stage, we estimate the potential structures in the target region $\mathbf{\Omega}$. To achieve this, we apply \emph{gPb Contour Detector}\cite{Pablo_11} to extract the edge distribution on the image:

\begin{equation}
I_{edge} = \sqrt{[\frac{\partial}{\partial x}(I \ast G_x)]^2 + [\frac{\partial}{\partial y}(I \ast G_y)]^2}
\end{equation}
where, $G_x$ and $G_y$ are the first derivative of Gaussian function with respect to $x$ and $y$ axis ($G_x = \frac{-x\cdot G(x, y)}{\sigma^2}$). After computing $I_{edge}$, most of the strong edges can be extracted via threshold suppression. Inspired by the level lines technique \cite{Anal_02}, the edges in $\Omega$ can be estimated by linking matching pairs of edges along the contour.

{\bf Structure Propagation:} After the structures are estimated, textures along the structures are synthesized and propagated into the target region $\mathbf{\Omega}$. We use Belief Propagation to identify optimal patches of texture from the source region $\Phi$ and copy and paste them to the structures in $\Omega$.

{\bf Remaining Part Filling:} After the structure propagation, the remaining unfilled regions in $\Omega$ are completed. We adopt the Criminisi's method  \cite{Criminisi_04}, where a priority-based patch filling scheme is used to render the remaining target region in an optimal order.

In the following subsections, we present the details of each step of the proposed algorithm. In particular, we give emphasis to the first two steps: structure estimation and structure propagation, which provide the most contribution of this proposed technique.

\subsection{Structure Estimation}
In this stage, we estimate the potential structures in $\Omega$ by finding all the possible edges. This procedure can be divided into two steps: Contour Detection in $\Phi$ and Curve Generation in $\Omega$.

\subsubsection{ Contour Detection in $\Phi$}
We first segment the region $\Phi$ by using \emph{gPb Contour Detector} \cite{Pablo_11}, which computes the oriented gradient signal $G(x, y, \theta)$ on the four channels of its transformed image: brightness, color \emph{a}, color \emph{b} and texture channel. $G(x, y,  \theta)$ is the gradient signal, where $(x, y)$ indicates the center location of the circle mask that is drawn on the image and $\theta$ indicates the orientation.  The \emph{gPb Detector} has two important components: \emph{mPb Edge Detector} and \emph{sPb Spectral Detector} \cite{Pablo_11}. According to the gradient ascent on F-measure, we apply a linear combination of \emph{mPb} and \emph{sPb} (factored by $\beta$ and $\gamma$):

\begin{equation}
gPb(x, y, \theta) = \beta\cdot mPb(x, y, \theta) + \gamma \cdot sPb(x, y, \theta)
\end{equation}

Thus a set of edges in $\Phi$ can be retrieved via \emph{gPb}. However, these edges are not in close form and have classification ambiguities. To solve this problem, we use the \emph{Oriented Watershed Transform}\cite{Pablo_11} and \emph{Oriented Watershed Transform}\cite{Arbel_06} (\emph{OWT-UCM}) algorithm to find the potential contours by segmenting the image into different regions. The output of \emph{OWT-UCM} is a set of different contours $\{C_i\}$ and their corresponding boundary strength levels $\{\mathcal{L}_i\}$.

\subsubsection{Curve Generation in $\Omega$}
After obtaining the contours $\{C_i\}$ from the above procedure, salient boundaries in $\Phi$ can be found by traversing $\{C_i\}$. Our method for generating the curves in $\Omega$ is based on the assumption: for the edges on the boundary in $\Phi$ that intersects with the $\partial\Omega$, it either ends inside $\Omega$ or passes through the missing region $\Omega$ and exits at another point of $\partial\Omega$. Below is our algorithm for identifying the curve segments in $\Omega$:

\begin{algorithm}[h]
\caption{
  {\bf {Identifying curve segments in $\Omega$}}}
\label{alg1}
\small
\begin{algorithmic}
\REQUIRE {Reconstruct missing curves segments in $\Omega$}
\ENSURE {The estimated curves provide smooth transitions between edges in $\Phi$}
\end{algorithmic}
\begin{algorithmic}[1]
  \STATE Initial $\mathbf{t} = 1.0$
  \STATE For $\mathbf{t} = \mathbf{t} - \Delta\mathbf{t}$
  \STATE \qquad if $ \exists e \in \{\mathcal{C}\}: E \cap \partial\Omega \neq \emptyset$
  \STATE  \qquad \qquad Insert $e$ into $\{E\}$
  \STATE End if $\mathbf{t} < \delta_T $
  \STATE Set $\mathbf{t} = \mathbf{t}_0$, retrieve all the contours in $\{C_i\}$ with $L_i > \mathbf{t}$
  \STATE Obtain $<\phi_{x1}, \phi_{x2}>$ for each $E_x$
  \STATE $\mathbf{DP}$ on $\{<\phi_{01}, \phi_{02}>, <\phi_{11}, \phi_{12}>, ...  \}$ to find optimal pairs
  \STATE According to the optimal pairs, retrieve all the corresponding edge-pairs: $\{(E_{x1}, E_{x_2}), (E_{x_3}, E_{x_4}), ...)\}$.
  \STATE Compute a transition curve $\mathcal{C}_{st}$ for each $\{(E_{s}, E_{t})\}$.
\end{algorithmic}
\end{algorithm}

In algorithm \ref{alg1}, it has three main parts: (a) collect all potential edges  $\{E_x\}$ in $\Phi$ that hits $\partial\Omega$; (b) identify  optimal edge pairs $\{(E_s, E_t)\}$ from $\{E_x\}$; (c) construct a curve $\mathcal{C}_{st}$ for each edge pair $(E_s, E_t)$.

{\bf Edges Collection:}  The output of \emph{OWT-UCM} are contours sets $\{\mathcal{C}_i\}$ and their corresponding boundary strength levels $\{\mathcal{L}_i\}$. Given different thresholds $\mathbf{t}$, one can remove those contours $\mathcal{C}$ with weak $\mathcal{L}$. Motivated by this, we use the \emph{Region-Split} scheme to gradually demerge the whole $\Phi$ into multiple sub-regions and extract those salient curves. This process is carried out on lines 1-9: at the beginning the whole region $\Phi$ is considered as one contour; then iteratively decrease $\mathbf{t}$ to let potential sub-contours $\{\mathcal{C}_i\}$ faint out according the boundary strength; Every time when any edges $e$ from the newly emerged contours $\{\mathcal{C}\}$ were detected of intersecting with $\partial\Omega$, they are put into the set $\{E\}$.

{\bf Optimal Edge Pairs:} the motivation of identifying edge pairs is based on the assumption if an edge is broken up by $\Omega$, there exists a pair of corresponding contour edges in $\Phi$ that intersect with $\partial\Omega$. To find the potential pairs  $\{(E_s, E_t)\}$ from the edge list $\{E_x\}$, we measure the corresponding enclosed regions similarities. The neighboring subregions $<\phi_{x_1}^{(s)}, \phi_{x_2}^{(s)}>$ which are partitioned by the edge $E_s$  are used to compare with the corresponding subregions $<\phi_{x_3}^{(t)}, \phi_{x_4}^{(t)}>$ of another edge $E_t$. This procedure is described on lines $7-9$ of the algorithm \ref{alg1}. For simplicity, the superscripts $(s)$ and $(t)$ are removed and the neighboring subregions $<\phi_{x_1}, \phi_{x_2}>$ are list in a sequential order. Each neighboring region is obtained by lowing down the threshold value $\mathbf{t}$ to faint out its contours.


 To compute the similarity between regions, we use the \emph{Jensen-Shannon divergence} \cite{Lamberti_08} method that works on the color histograms:

\begin{equation}
d(H_1, H_2) = \sum_{i = 1}^n \{H_1^i\cdot log\frac{2\cdot H_1^i}{H_1^i + H_2^i} + H_2^i\cdot log\frac{2\cdot H_2^i}{H_2^i + H_1^i} \}
\end{equation}

where $H_1$ and $H_2$ are the histograms of the two regions $\phi_{1}, \phi_{2}$; $i$ indicates the index of histogram bin. For any two edge $(E_{s}, E_{t})$, the similarity between them can be expressed as:

\begin{equation}
M(E_s, E_t) = \frac{||L_s - L_t||}{L_{max}}\cdot\min\{d(H_{si}, H_{ti}) + d(H_{sj}, H_{tj})\}
\end{equation}

$i$ and $j$ are the exclusive numbers in $\{1, 2\}$, where 1 and 2 represent the indices of the two neighboring regions in $\Phi$ around a particular edge. The $L_{max}$ is the max value of the two comparing edges' strength levels. The first multiplier is a penalty term for big difference between the  strength levels of the two edges. To find the optimal pairs among the edge list, dynamic programming is used to minimize the global distance: $\sum_{s, t}M(E_s, E_t)$, where $s \neq t$ and $s, t \in \{0, 1, ..., size(\{E_i\})\}$. To enhance the accuracy, a maximum constraint is used to limit the regions' difference: $d(H_1, H_2) < \delta_H$. If the individual distance is bigger than the pre-specified threshold $\delta_H$, the corresponding region matching is not considered. In this way, it ensures if there are no similar  edges existed, no matching pairs would be identified.

{\bf Generate Curves for each $(E_s, E_t)$ :} we adopt the idea of fitting the clothoid segments with polyline stoke data first before generating a curve \cite{McCrae_09}. Initially, a series of discrete points along the two edges $E_s$ and $E_t$ are selected, denoted as $\{p_{s0}, p_{s1}, ..., p_{sn}, p_{t0}, p_{t1}, ..., p_{tm}\}$. These points have a distance with each other by a pre-specified value $\Delta_d$. For any three adjacent points $\{p_{i-1}, p_{i}, p_{i+1}\}$, the corresponding curvature $k_i$ could be computed according to \cite{Mullinex_07}:

\begin{equation}
k_i = \frac{2\cdot det(p_i - p_{i - 1}, p_{i + 1} - p_i)}{||p_i - p_{i-1}||\cdot||p_{i + 1} - p_i||\cdot||p_{i+1} - p_{i -1}||}
\end{equation}

Combining the above curvature factors, a sequence of polyline are used to fit these points. The polylines are expected to have a possibly small number of line segments while preserving the minimal distance against the original data.  Dynamic programming is used to find the most satisfied polyline sequence by giving a penalty for each additional line segment. A set of clothoid segments can be derived corresponding to each line segment. After a series rotations and translations over the clothoid, a final curve $\mathcal{C}$  is obtained by connecting each adjacent pair with $\mathbf{G}^2$ continuity \cite{McCrae_09}.

\subsection{ Structure Propagation}
After the potential curves are generated in $\Omega$, a set of texture patches, denoted as $\{\Psi_0, \Psi_1, ... \}$, need to be found from the remaining region $\Phi$ and placed along the estimated curves by overlapping with each other with a certain proportion. Similar to the Sun's method \cite{sun_05}, an energy minimization based Belief Propagation(BP) framework is developed. We give different definitions for the energy minimization and passing messages, the details of which can be found in algorithm \ref{alg3}.

\begin{algorithm}[h]
\caption{
  Belief Propagation for Structure Completion }
\label{alg3}
\small
\begin{algorithmic}
\REQUIRE {Render each patch $\Psi_i$ along the estimated structures  in $\Omega$ }
\ENSURE {Find the best matching patches while ensuring texture conherence}
\end{algorithmic}
\begin{algorithmic}[1]
  \STATE For each curve $\mathcal{C}$ in $\Omega$, define a series of \emph{anchor points} on it, $\{\mathbf{a}_i, |i = 1\rightarrow n\}$
  \STATE Collect exemplar-texture patches $\{\hat{\Psi}_{t_i}\}$ in $\Phi$, where $t_i\in[1, m]$
  \STATE Setup a factor graph $\mathcal{G} = \{\mathcal{V}, \mathcal{E}\}$ based on $\{\mathcal{C}\}$ and $\{\mathbf{a}_i\}$
  \STATE Defining the energy function $\mathbf{E}$ for each $\mathbf{a}_i$: $\mathbf{E}_i(t_i)$, where $t_i$ is the index in $[1, M]$
  \STATE Defining the message function $\mathbf{M}_{ij}$ for each edge $\mathcal{E}$ in $\mathcal{G}$, with initial value $\mathbf{M}_{ij} \leftarrow 0$
  \STATE Iteratively update all the messages $\mathbf{M}_{ij}$ passed between $\{\mathbf{a}_i\}$
  \STATE \qquad $\mathbf{M}_{ij} \leftarrow \min_{a_i}\{ \mathbf{E}_i(t_i) + \mathbf{E}_{ij}(t_i, t_j) + \sum_{k\in \mathcal{N}(i), k\neq j}\mathbf{M}_{ki}\}$
  \STATE end until $\Delta \mathbf{M}_{ij} < \delta$, $\forall i, j$ (by Convergence)
  \STATE Assign the best matching texture patch from $\{\hat{\Psi_t}\}$ for each $\mathbf{a}_i$ that $\arg \min_{[T, R] }\{\sum_{i\in \mathcal{V}}\mathbf{E}_i(t_i) + \sum_{(i, j)\in \mathcal{E}}\mathbf{E}_{ij}(t_i, t_j)\}$, where $T$ and $R$ represent the translation and orientation of the patch $\{\hat{\Psi}_{t_i}\}$
\end{algorithmic}
\end{algorithm}

In our algorithm, the \emph{anchor points} are evenly distributed along the curves with an equal distance from each other $\Delta d$. These points represent the center where the patches $\{\Psi_i\}$ ($l\times l$) are placed. In practice, we define $\Delta{d} = \frac{1}{4}\cdot l$. The $\{\hat{\Psi}_t\}$ is the source texture patches in $\Phi$. They are chosen on from the neighborhood around $\mathbf{\partial\Omega}$. According to the Markov Random Field definition, each $\mathbf{a}_i$ is considered as a vertex $\mathcal{V}_i$ and $\mathcal{E}_{ij} = \mathbf{a}_i\mathbf{a}_j$ represents a edge between two neighboring nodes $i$ and $j$.

Among the traditional exemplar-based methods, when copy a texture patch from the source region $\Phi$ to the target region $\Omega$, each $\Psi_i$ have the same orientation as $\hat{\Psi}_{t_i}$, which limits the varieties of the texture synthesis.  Noticing that different patch orientations could produce different results, we introduce a scheme called \emph{Adaptive Patch} by defining a new configuration for the energy metric $\mathbf{E}$ and message $\mathbf{M}$.

Intuitively, the node energy $\mathbf{E}_i(t_i)$ can be defined as the \emph{Sum of Square Difference}(SSD) by comparing the known pixels in each patch $\Psi_i$ with the candidate patch in $\hat{\Psi}_{t_i}$. But this could limit the direction changes of the salient structure. So instead of using SSD on the two comparing patches, rotation transformation is performed to the candidate patch before computing the SSD. Mathematically, $\mathbf{E}_i(t_i)$ can be formulated as:

\begin{equation}
\mathbf{E}_i(t_i) = \alpha_\lambda\cdot P\cdot\sum||\Psi_i - \dot{R}(\theta)\cdot\hat{\Psi}_{t_i}||_{\lambda}^2
\end{equation}
where $\dot{R}$ represents the 2D rotation matrix with an input angle parameter $\theta$ along the orthogonal vector that is perpendicular to the image plane. Since the size of a patch is usually small, the rotation angle $\theta$ can be specified with an arbitrary number of values. In our experiment, it is defined as $\theta \in \{0, \pm\frac{\pi}{4}, \pm\frac{\pi}{2}, \pi\}$. Parameter $\lambda$ represents the number of known pixels in $\Psi_i$ that overlap with the rotated patch $\hat{\Psi}_{t_i}$. $P$ is a penalty term that discourage the candidate patches with smaller proportion of overlapping pixels with the neighboring patches. Here, we define $P$ as $P = \frac{\lambda}{l^2}$ ($l$ is the length of $\Psi$). $\alpha_\lambda$ is the corresponding normalization factor.

In a similar way, the energy $\mathbf{E}_{ij}(t_i, t_j)$ on each edge $\mathcal{E}_{ij}$ can be expressed as:

\begin{equation}
\mathbf{E}_{ij}(t_i, t_j) = \alpha_{\lambda}\cdot P\cdot\sum||\Psi_i(t_i, \theta_{t_i}) - \Psi_j(t_j, \theta_{t_j})||_\lambda^2
\end{equation}
where $i$ and $j$ are the corresponding indices of the two adjacent patches in $\Omega$. The two parameters for $\Psi_i$ indicate the index and rotation for the source patches in $\{\hat{\Psi}_{t_i}\}$. We adopt a similar message passing scheme as \cite{sun_05} that message $M_{ij}$ passes by patches $\Psi_i$ is defined as:

\begin{equation}
M_{ij} = \mathbf{E}_i(t_i) + \mathbf{E}_{ij}(t_i, t_j)
\end{equation}

Through iterative message passing on the MRF graph to minimize the global energy, an optimal configuration of $\{t_i\}$ for the patches in $\{\Psi_i\}$ can be obtained. The optimal matching patch index $\hat{t}_i$ is defined as:

\begin{equation}
\hat{t}_i = \arg \min_{t_i}\{\mathbf{E}_i(t_i) + \sum_k M_{ki}\}
\end{equation}
Where $k$ is the index of one of the neighbors of the patch $\Psi_i$: $k \in\mathcal{N}(i)$.  To compute an minimum energy cost, dynamic programming is used: at each step, different states of $\hat{\Psi}_{t_i}$ can be chosen. The edge $\mathcal{E}_{ij}$ represents the transition cost from  the state of $\hat{\Psi}_{t_i}$ at step $i$ to state of $\hat{\Psi}_{t_j}$ at step $j$. Starting from $i = 0$, an optimal solution is achieved by minimizing the total energy $\xi_i(t_i)$:

\begin{equation}
\xi_i(t_i) = \mathbf{E}_i(t_i) + min\{\mathbf{E}_{ij}(t_i, t_j) + \xi_{i-1}(t_{i-1})\}
\end{equation}
 where $\xi_i(t_i)$ represents a set of different total energy values at the current step $i$. In the cases of multiple intersections between curves $\mathcal{C}$, we adopted the idea of Sun's method \cite{sun_05}, where readers can refer to for further details.

\subsection{Remaining Part Filling}
After the structure curves are generated in $\Omega$, we fill the remaining regions by using the exemplar-based approach in \cite{Criminisi_04}, where patches are copied from the source region $\Phi$ to the filling region $\Omega$ in a priority order. The priority is determined by the extracted edges in $\Phi$ that intersect with $\mathbf{\partial\Omega}$. To ensure the propagated structures in $\Omega$ maintain the same orientation as in $\Phi$,  higher priorities of texture synthesis are given to those patches that lie on the continuation of stronger edges in $\Phi$.

According to Criminisi's algorithm \cite{Criminisi_04}, each pixel on a image has a confidence value and color value. The color value can be empty if it is in the unfilled region $\Omega$. For a given patch $\Psi_\mathbf{p}$ at a point $\mathbf{p}$, its priority is defined as: \emph{priority}({$\mathbf{p}$}) = $C(\mathbf{p})\cdot D(\mathbf{p})$, where $C(\mathbf{p})$ and $D(\mathbf{p}$ are the confidence map and data term that are define as:

\begin{equation}
C(p) = \frac{\sum_{\mathbf{p}\in \Phi_\mathbf{p} \bigcap (\mathcal{I} - \Omega)}\cdot C(\mathbf{q})}{|\Psi_\mathbf{p}|}
\end{equation}

and

\begin{equation}
D(p) = \frac{|\Delta I_\mathbf{p}^x\cdot \mathbf{n^\perp_p}|}{\alpha}
\end{equation}
where $\mathbf{q}$ represents the surrounding pixels of $\mathbf{p}$ in the patch $\Psi_\mathbf{p}$. $|\Psi_\mathbf{p}|$ is the area of the patch $\Psi_\mathbf{p}$. The variable $\mathbf{n_p}$ is a unit orthogonal vector that is perpendicular to the boundary $\mathbf{\partial\Omega}$ on the point $\mathbf{p}$. The normalization factor $\alpha $ is set as 255 as all the pixels are in the range $[0, 255]$ for each color channel. So in such a way,  the priority for each pixel can be computed. For further details, we refer readers to the Criminisi's algorithm \cite{Criminisi_04} for more explanations.

\section{Experiments}
\label{sec:exp}

In our experiments, to evaluate our algorithm, different styles of images are tested from natural scenes to indoor environment that has strict structure. Our algorithm obtains satisfactory results in terms of texture coherence and structural consistency.  The algorithm is implemented in C++ code with OpenCV library. All the images results are generated on a dual-core PC with CPU 2.13GHz and Memory 2G. For the images with the regular resolution $640 \times 480$, the average computation cost is about 52 seconds.

Figure 2 demonstrates the advantage of our method by preserving the scene structure after removing the occluded object. The original image data can be find publicly at the website \footnote{http://www.cc.gatech.edu/~sooraj/inpainting/}. Figure 2 shows the target region (the bungee jumper) for removal marked in green color. Figure 3, 4 are the image reconstruction results by the Criminisi's and our methods respectively. One can notice the roof area in figure 5 is broken by the grass which introduces noticeable artifact, while the corresponding part remains intact in our result. Furthermore, in contrast to the Criminisi's method, the lake boundary is naturally recovered thanks to our structure estimation procedure, as shown in 6. In terms of the time performance for the original image of $205 \times 307$ pixels, our method performed $10.5$ seconds on the computer of dual-core PC with CPU 2.13GHz and 2GB of RAM, to be compared with $18$ seconds of Criminisi's on a 2.5 GHz Pentium IV with 1 GB of RAM.

Another existing work we choose to compare with is the Sun's method, which also aims to preserve the original structure in the recovered image. However, the difference is that Sun's method requires manual intervention during the completion process. The potential structure in the target region needs to be manually labeled by the designer, which can vary according to individuals. Figure 7 demonstrates a comparison between Sun's and our methods. In the original image, the car is considered as the target object for removal, which is marked in black color in figure 8.   In figures 9, the potential structures in the target region are labeled by \cite{sun_05} and automatically estimated by our method, which produce different results, as shown in figures 10, 11. To compare the computation speed, our methods performed 51.7 seconds to process this image ($640\times 457)$, in contrast with the Sun's fewer than 3 seconds for each curve propagation (3 curves in total) and 2 to 20 seconds for each subregion (4 subregions in total) on a 2.8 GHz PC. Moreover, we save the potential labor work on specifying the missing structures by the user.

%
%

To further demonstrate the performance, a set of images are used for scene recovery: ranging from indoor environment to natural scenes.  Figure 12 shows the case of indoor environment, where highly structural patterns often present, such as the furniture, windows, walls. In figure 12, the green bottle on the office partition is successfully removed while preserving the remaining structure. In this example, five pairs of edges are identified and connected by the corresponding curves that are generated in the occluded region $\Omega$. Guided by the estimated structure, plausible texture information is synthesized to form a smooth intensity transition across the occluded region with little artifact.

For the outdoor environment, as there are fewer straight lines or repeating patterns existing in the natural world, the algorithm should provide the flexibility to generate irregular structures. Figure 13 and 14 show the results of removing trees in the nature scenes. Several curves are inferred by matching the broken edges along $\partial\Omega$ and maximizing the continuity. We can notice the three layers of the scene (sky, background trees, and grass land) are well completed. In Figure 15, it shows a case that a perching bird is removed from the tree. Our structure estimation successfully completes the tree branch with smooth geometric and texture transitions. Without such structure guidance, those traditional exemplar-texture based methods can easily produce noticeable artifacts. For example, multiple tree branches may be generated as the in-painting process and directions largely rely on the matching patches.

\section{Conclusion}
\label{sec:con}

In this paper, we present a novel approach for foreground objects removal while ensure structure coherence and texture consistency. The core of our approach is to use structure as a guidance to complete the remaining scene, which demonstrates its accuracy and consistency.  This work would benefit a wide range of applications, from digital image restoration (e.g. scratch recovery) to privacy protection (e.g. remove people from the scene). In particular, this technique can be promising for the online massive collections of imagery, such as photo localization and scene reconstructions. By removing foreground objects, the matching accuracy can be dramatically improved as the corresponding features are only extracted from the static scene rather than those moving objects. Furthermore it can generate more realistic views because the foreground pixels are not involved in any image transformation and geometric estimation.

As one direction of our future work, we will apply this object removal technique to scene reconstruction applications that can generate virtual views or reconstruct the 3D data from a set of images. Multiple images can give more cues of the potential structure and texture in the target region. For example, through corresponding features among different images, intrinsic and extrinsic parameters can be estimated. Then the structure and texture information can be mapped from one image to another image. So for a particular target region for completion, multiple sources (from different images) can contribute the estimation. As such, Our current algorithm needs to be modified adaptively to take the advantage of the extra information.  An optimization framework could be established to identify optimal structures and textures to fill the target region.

\bibliography{ref}

\section*{Biography}

\medskip
\noindent
{\bf Ju Shen} is an Assistant Professor from the Department of Computer Science, University of Dayton (UD),
Dayton, Ohio, USA. He received his Ph.D. degree from University of Kentucky, Lexington, KY, in USA and his M.Sc degree
from University of Birmingham, Birmingham, United Kingdom. His work spans a number of different areas in computer vision, graphics,
multimedia, and image processing. \\
\vspace*{0.5cm}

\medskip
\noindent
{\bf Jianjun Yang} received his B.S. degree and his first M.S. degree in Computer Science in China, his second M.S. degree in Computer Science during his doctoral study from University of Kentucky, USA in May 2009, and his Ph.D degree in Computer Science from University of Kentucky, USA in 2011. He is currently an Assistant Professor in the Department of Computer Science and Information Systems at the University of North Georgia. His research interest includes wireless networking, computer networks, and image processing.. \\
\vspace*{0.5cm}

\medskip
\noindent
{\bf Sami Taha Abu Sneineh} is currently an assistant professor at Palestine Polytechnic University (PPU). He received his Ph.D. degree in May 2013 in Computer Vision at the University of Kentucky (UK) under the supervision of Dr. Brent Seales.  Sami earned a B.Sc in Electrical Engineering from Palestine Polytechnic University in 2001 and M.Sc in Computer Science from Maharishi University of Management in 2006. His research focus is on computer vision in minimally invasive surgery to improve the performance assessment. Sami has worked for three years as consultant software engineer at Lexmark Inc. and four years as consultant programmer at IBM before joining UK. He started his teaching experience as TA in the computer science department. He received a certificate and an award in college teaching and learning in 2012.  His goal is to contribute in improving the higher education system and raise the education standards. \\
\vspace*{0.5cm}

\medskip
\noindent
{\bf Bryson Payne} received his B.S. degree in Mathematics from North Georgia College \& State University, USA in 1997, his M.S. degree Mathematics from North Georgia College \& State University, USA in 1999, and his Ph.D degree in computer science from Georgia State University, USA in 2004. He was the (CIO )Chief Information Officer in the North Georgia College \& State University(former University of North Georgia). He is currently an Associate Professor in the Department of Computer Science and Information Systems at the University of North Georgia. He is also a CISSP(Certified Information Systems Security Professional). His research interest includes image processing, web application and communications.\\
\vspace*{0.5cm}

\medskip
\noindent
{\bf Markus Hitz} received his B.S. degree and M.S. degree of Computer Science in Europe and his Ph.D degree of Computer Science in USA. He is currently a Professor and Acting Head in the Department of Computer Science and Information Systems at the University of North Georgia. His research interest includes networking, simulation and communications.

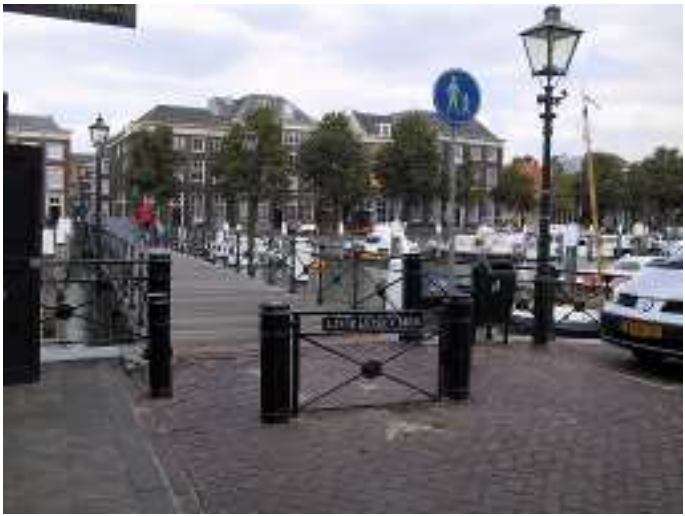
\end{document}